\documentclass[letterpaper]{article}
\usepackage[preprint]{aaai2027}
\usepackage[hyphens]{url}
\usepackage{graphicx}
\usepackage{natbib}
\usepackage{caption}
\usepackage{amsmath,amssymb,amsfonts}
\usepackage{booktabs}
\usepackage{multirow}
\usepackage[protrusion=false]{microtype}
\usepackage{enumitem}

\graphicspath{{figs/}}

\newcommand{\psig}{P_{\mathrm{sig}}}
\newcommand{\method}{CANOPY}

\title{Explore More, Drift Less: Outcome-Only Reinforcement Learning\\ Can Suffice for Long-Horizon Interactive Agents}
\author{
    Liming Pu,
    Xiaoxia Li,
    Yifu Liu,
    Teng Cao,
    Bin Yang
}
\affiliations{}

\begin{document}
\maketitle

\begin{abstract}
Reinforcement learning is a natural way to post-train LLM agents for long-horizon interactive tasks judged only by end-of-task verification, yet a shared belief holds that outcome-only RL soon hits a ceiling on small open models. Recent work therefore compensates around the training with denser rewards, SFT priors, skill libraries, curated memory, or multi-agent orchestration. We argue the ceiling is an artifact of two failures of common practice. \emph{Signal starvation}: group-relative RL with sparse outcome-only rewards yields a gradient only when a task's rollout group mixes successes and failures, so under-scaled exploration silences exactly the hardest, most instructive tasks. \emph{Policy drift}: squeezing many updates out of a small task pool degrades the policy itself, as an unanchored objective lets the sampling distribution collapse exactly when saturation has already made informative groups rare. We present \method{} (Coverage-ANchored On-PolicY RL), a minimalist protocol attacking both directly: scale same-task exploration until the natural signal reappears, keep every update on-policy, KL-anchored, and confined to the agent's own action tokens, then cash in an enlarged interaction budget at test time. On AppWorld, a long-horizon interactive coding benchmark, a Qwen3-14B policy trained with \method{} through environment interaction alone---without task-specific supervision, auxiliary credit signals, or elaborate agent scaffolding---topped the public leaderboard (Feb.~2026; Test-Normal TGC 86.9, Test-Challenge 67.6), and the same design principles lift Qwen3.5-9B on SWE-bench Verified by 16.6 points. Agentic RL alone thus internalizes long-horizon capability directly into a small open model; we plan to release the complete training stack at \url{https://github.com/AlibabaResearch/SignalCoverageRL}.
\end{abstract}

\section{Introduction}
\label{sec:intro}

LLM-driven agents are moving rapidly from demonstrations into daily work. Coding-centric agents now automate software engineering, office workflows, and everyday application operation, almost always through one architecture: an engineered harness wrapped around a frontier, usually closed, model. Post-training a small open model into a domain specialist is one route worth weighing alongside this: let reinforcement learning internalize a well-defined domain's interaction skills into the weights, and deploy a single lightweight policy with no external machinery. In AppWorld~\citep{trivedi2024appworld}, a long-horizon benchmark of everyday digital-application tasks, an agent iteratively writes and executes Python against a live environment, taking dozens of think--code--execute--observe turns before a task is judged by held-out, state-based unit tests.

\begin{figure}[t]
\centering
\includegraphics[width=0.98\columnwidth]{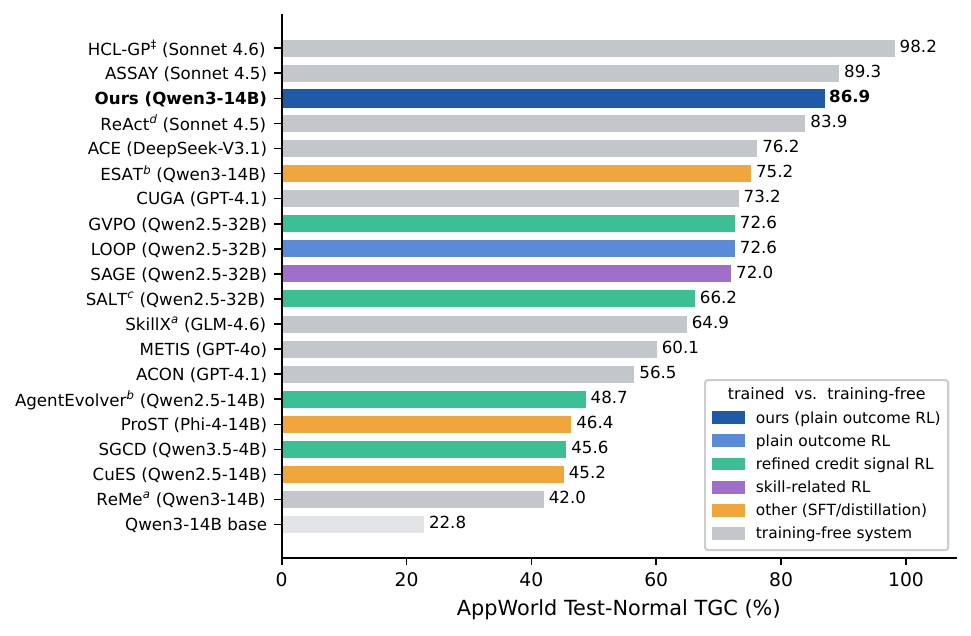}
\caption{\textbf{AppWorld Test-Normal TGC by method family} (official leaderboard~\citep{appworldleaderboard} or paper-reported; all entries mean@1 unless marked $^{a}$mean@4, $^{b}$mean@8, $^{c}$mean@3; $^{d}$reported by \citet{assay2026}; $^{\ddagger}$non-standard joint-scenario protocol with test-time debugging). Color encodes method family: four trained-policy sub-classes plus gray for training-free systems. Among trained policies, \method{} (bold, darkest blue) leads on one of the smallest backbones here. Entries differ in backbone, harness, and sampling protocol, so the ordering places families rather than ranking systems on one axis.}
\label{fig:landscape}
\end{figure}

Agentic reinforcement learning---a policy interacting autonomously with an environment and learning from its outcome feedback---is now the dominant approach to building such specialists, and AppWorld has become a proving ground for it. Long horizons and outcome-only verification make the setting hard, and most work on it responds by \emph{compensating around the policy} rather than strengthening the RL itself, each compensation paired with a finding that reads as evidence for a real ceiling: denser or step-level rewards, after the hardest tasks were measured as harmful~\citep{chen2025loop,gvpo2026,salt2026}; an SFT cold start, after skipping it was shown to collapse scores~\citep{sage2025}; skill libraries, curated memory, or multi-agent orchestration anchored to frontier closed models, after plain interaction at inference was reported to yield little for untrained agents~\citep{hclgp2025,generalagentbench2026}. RL has additionally been argued to narrow rather than expand the base model's capability boundary~\citep{szot2026sge,yue2025limit}.

Is this ceiling real, or an artifact of how the policy was trained? We argue the training trajectory distribution is distorted relative to what the policy must produce at test time, and that this distortion, not any limit of outcome-only RL, is the cause. Shortened horizons, filtered-out hard tasks, and few rollouts per task deprive the policy of complete, self-generated, error-recovering trajectories; stale off-policy reuse trains it on behavior no longer its own; substituted reward signals carry their own errors. Under group-relative policy optimization~\citep{shao2024deepseekmath}, this distortion collapses the learning signal itself, through two failures we name and then dismantle one at a time.

\emph{Failure 1: signal starvation.} A group-relative estimator yields useful gradient only when a task's rollout group contains both successes and failures. For per-rollout success rate $p$ and group size $n$, the probability of drawing such a group is $\psig(p,n)=1-p^n-(1-p)^n$. With the small groups used by prior RL work on this benchmark ($n\le 8$;~\citealp{chen2025loop,gvpo2026,sage2025}) and low $p$ on hard tasks, most groups are degenerate---zero outcome-reward gradient---and the occasional isolated success is amplified by standard-deviation normalization into a high-variance spike. This is exactly why prior work measured hard tasks as ``harmful'' and reached for dense or step-level signal: \emph{a compensation for under-exploration, not a property of the data}. Once exploration is scaled to restore coverage, the same hard tasks flip from poison to the most valuable data.

\emph{Failure 2: policy drift.} An interactive environment exposes a limited pool of verifiable tasks, so training must revisit them many times. Under such repetition the sampling distribution can collapse: without an anchor, entropy decays and exploration dies exactly when saturation already makes informative groups rare---the distortion of Failure 1, now acting on the policy rather than on a single gradient step. The visible symptom is late-training instability. Common defenses---early stopping, an SFT prior~\citep{sage2025,prost2025}, conservative horizons~\citep{chen2025loop}---avoid it by capping exactly the capability RL was meant to grow; the cap is then read as a ceiling of RL itself.

\paragraph{A simple, minimal protocol.}
\method{} (Coverage-ANchored On-PolicY RL) attacks both failures under one principle, using well-understood ingredients throughout: manufacture the signal, then keep it trustworthy. \textbf{Explore more} answers starvation with large same-task groups and uncapped per-turn generation, keeping the hardest tasks in the pool. \textbf{Drift less} answers drift with a light KL anchor, strictly on-policy updates, and a token-level loss over action tokens only. \textbf{Realize at test time} transfers the trained policy to an enlarged interaction budget. Trained this way, a Qwen3-14B policy lifted its base by more than 50 TGC points and reached the top of the AppWorld leaderboard (Feb.~2026; Test-Normal TGC 86.9, Test-Challenge 67.6) on the lightest configuration among the trained policies we compare against (Figure~\ref{fig:landscape}, Table~\ref{tab:main}); the same design principles lift a Qwen3.5-9B policy by 16.6 points on SWE-bench Verified. Our contributions:

\begin{itemize}[leftmargin=1.2em,itemsep=2pt,topsep=3pt,parsep=0pt]
\item \textbf{A diagnosis.} We trace the apparent ceiling to a distorted training-trajectory distribution and isolate two mechanisms---signal starvation and policy drift---that reconcile findings reading as contradictory across papers: the hard tasks reported as harmful are only starved of signal, and become the last gradient source once coverage is restored.

\item \textbf{A minimal protocol.} \method{} pairs each mechanism with one well-understood ingredient, changing no optimizer and adding no auxiliary module, and carries over from application operation to real-repository software repair under the same design principles. We report it with its full configuration, per-split metrics under one protocol, and six ablations pricing each ingredient.

\item \textbf{A position.} Plain agentic RL on small open models is not the exhausted direction. A single open 14B policy, trained by interaction alone with no SFT prior, skill library, or orchestration, holds its own against far heavier inference-time systems on stronger backbones, and expands---not merely resharpens---its base model's boundary. The field's turn toward harness engineering answers trainable ceilings that our results place higher than reported.
\end{itemize}

\section{Related Work}
\label{sec:related}

\paragraph{Coding agents and agentic RL.}
Interactive coding agents---systems that autonomously plan, write code, execute it, and act on the result over many turns---have become a mainstream way to deploy LLMs, spanning software engineering~\citep{yang2024sweagent,swerebench2025}, web and research tasks~\citep{mirothinker2025}, and everyday application operation, where AppWorld~\citep{trivedi2024appworld} is the canonical long-horizon benchmark. Outcome-verified RL, first matured on single-turn mathematics and code~\citep{deepseekr1,shao2024deepseekmath}, is now a core post-training technique for these multi-turn agents. Our study lives at this intersection: application operation as the primary testbed and software repair as the transfer domain.

\paragraph{RL algorithms and their failure modes.}
The PPO~\citep{schulman2017ppo} and GRPO~\citep{shao2024deepseekmath} family dominates agent post-training, GRPO especially: a group of rollouts replaces the learned critic, which makes it simple to train, widely adopted, and the base of many variants---as is the leave-one-out estimator RLOO~\citep{ahmadian2024rloo}. These variants keep the group-relative structure while adjusting the objective---bias analysis (Dr.~GRPO;~\citealp{liu2025drgrpo}), sequence-level importance ratios (GSPO;~\citealp{gspo2025}), turn-level grouping (GiGPO;~\citealp{gigpo2025}), graph-global credit assignment across sampled trajectories (G2PO;~\citealp{g2po2026}), dynamic sampling of uninformative prompts (DAPO;~\citealp{yu2025dapo})---while PPO-family training remains in use for large agentic models~\citep{glm45,sropo2026}. On failure modes, zero-advantage groups under identical rewards have been formalized as advantage collapse with a virtual-sample fix~\citep{advcollapse2026}; a recipe study concludes small models need staged rewards~\citep{demystify2026}. These works fix the estimator or the reward while leaving exploration as given; we instead show that scaling exploration removes the need for such fixes.

\defcitealias{sgcd2026}{Ding et al. 2026}
\paragraph{Policy training on AppWorld.}
Figure~\ref{fig:landscape} groups policy training into four routes.
\emph{(1) Plain outcome RL}: LOOP~\citep{chen2025loop} uses leave-one-out PPO and pass-fraction rewards, and drops the hardest tier as harmful---the starvation regime our analysis predicts; SeeUPO~\citep{seeupo2026} derives sequence-level updates.
\emph{(2) Refined credit signals}: SALT~\citep{salt2026}, GVPO~\citep{gvpo2026}, and AgentEvolver~\citep{agentevolver2025} respectively use trajectory-graph redistribution, execution-process signals, and self-generated curricula with step-level judges; SGCD~\citepalias{sgcd2026} reweights GRPO using a training-only external-LLM reference built from mixed-outcome sibling rollouts.
\emph{(3) Skill-related RL} uses skills in training but differs at deployment. Skill-SD~\citep{skillsd2026} self-distills from a skill-conditioned teacher to a plain-prompt student; SAGE~\citep{sage2025} trains and retains a skill library across scenario tasks, using sequential rollouts, a skill-integrated reward, and expert-trajectory SFT.
\emph{(4) Other training}: ProST~\citep{prost2025} progressively fine-tunes role-specialized small agents on frontier-model trajectories; CuES~\citep{cues2025} synthesizes executable, environment-grounded RL tasks; ESAT~\citep{esat2026} builds SFT data with generated tasks, teacher trajectories, and simulated API responses, without executing AppWorld. Each route compensates for starvation with denser signal, priors, or auxiliary structure; none removes it, and all keep the exploration budget small.

\paragraph{Training-free AppWorld systems.}
A parallel line engineers inference-time systems around fixed frontier models: multi-agent orchestration (CUGA;~\citealp{cuga2025}), hierarchical policy-decomposition reuse with test-time debugging (HCL-GP;~\citealp{hclgp2025}), causal measurement and per-task masking of natural-language skills (ASSAY;~\citealp{assay2026}), automatically constructed hierarchical skill knowledge bases transferred across agents (SkillX;~\citealp{skillx2026}), evolving playbooks and procedural memory~\citep{ace2025,reme2025,metis2026}, and context compression~\citep{acon2025}. These inherit dependence on a strong backbone (open or closed), per-scenario engineering cost, and runtime complexity, and the capability never enters the weights (Section~\ref{sec:main}). On test-time scaling, \citet{generalagentbench2026} report that more turns yield little for untrained generic agents; others sample many rollouts and select with a verifier, or generate skills first~\citep{sage2025}. We instead enlarge the trained policy's context and turn budget.

\section{The \method{} Protocol}
\label{sec:core}

We develop our diagnosis and protocol together, one failure at a time: each subsection names a mechanism that can stall agentic RL, then the practice that answers it.

\subsection{Preliminaries: The Agentic RL Loop}
\label{sec:setup-rl}

An agentic RL loop couples a policy to an environment: for a task prompt $q$, the policy $\pi_\theta$ proposes an action, the environment executes it and returns feedback, and this repeats until the policy terminates or the episode exhausts its turn or context budget. We write an episode as a trajectory $o=(a_1,e_1,a_2,e_2,\dots)$, interleaving \emph{action tokens} $a_t$---the agent's thinking and code---and \emph{environment tokens} $e_t$, the execution feedback (a sandboxed Python interpreter here, a shell in Section~\ref{sec:swe}). A held-out unit-test suite $U(q)$ judges the episode. Following the outcome-reward formulation of GVPO~\citep{gvpo2026}, let $u_j\in U(q)$, $j=1,\dots,M$ be the $M$ unit tests for $q$ and $\mathrm{pass}(u_j, o_i)\in\{0,1\}$ their result on trajectory $o_i$. The \emph{dense} pass-fraction reward reports the fraction passed,
\begin{equation}
r_i^{\mathrm{dense}} \;=\; \frac{1}{M}\sum_{j=1}^{M} \mathrm{pass}(u_j, o_i) \;\in\; [0,1],
\label{eq:dense}
\end{equation}
giving partial credit to a trajectory that passes some tests even if its overall approach is wrong. The \emph{sparse} reward instead credits only a fully correct trajectory,
\begin{equation}
r_i^{\mathrm{sparse}} \;=\; \mathbf{1}\!\left[\textstyle\sum_{j=1}^{M} \mathrm{pass}(u_j, o_i) = M\right] \;\in\; \{0,1\}.
\label{eq:sparse}
\end{equation}
Section~\ref{sec:drift} revisits the choice. Group-relative policy optimization~\citep{shao2024deepseekmath} samples a group of $n$ trajectories $\{o_i\}_{i=1}^{n}$ for the same task under the current policy and turns the reward $r_i$ into a standardized within-group advantage,
\begin{equation}
\hat A_i \;=\; \frac{r_i - \mathrm{mean}(r_1,\dots,r_n)}{\mathrm{std}(r_1,\dots,r_n)} .
\label{eq:adv}
\end{equation}
One iteration then closes as follows: roll out $n$ trajectories per task for a batch of tasks, score them with $U(q)$, convert rewards to advantages by Equation~\ref{eq:adv}, take a gradient step on the policy that produced them (loss in Section~\ref{sec:drift}), and let the updated policy sample the next batch. This sampling-and-update pair, with no learned critic, is what the rest of the section builds on.

\begin{figure}[h]
\centering
\includegraphics[width=0.98\columnwidth]{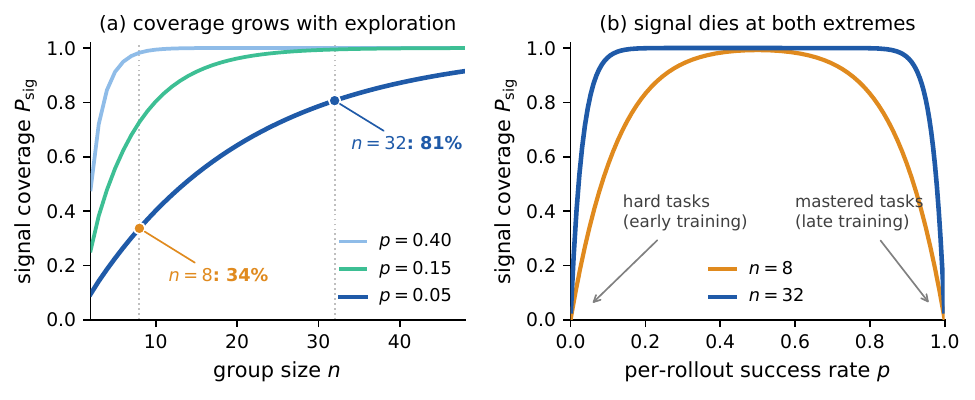}
\caption{\textbf{Signal coverage under a sparse reward.} (a) Coverage $\psig$ vs.\ group size $n$: for a hard task ($p{=}0.05$) it rises from 34\% at $n{=}8$ to 81\% at $n{=}32$, then flattens. (b) Coverage vs.\ per-rollout success rate $p$ at fixed $n$: signal collapses at both extremes---hard tasks early in training, mastered tasks late.}
\label{fig:psignal}
\end{figure}

\subsection{Signal Starvation, and Explore More}
\label{sec:starvation}

\paragraph{Diagnosis: the coverage mechanism.}
With the sparse reward of Equation~\ref{eq:sparse}, Equation~\ref{eq:adv} is non-zero only when a group contains at least one success \emph{and} one failure; if all $n$ rollouts fail (or all succeed), every advantage is zero and the task contributes no gradient at that step. Call a group with mixed outcomes \emph{informative}, and the probability of drawing one the task's \emph{signal coverage}. If rollouts succeed independently with probability $p$,
\begin{equation}
\psig(p, n) \;=\; 1 - p^n - (1-p)^n .
\label{eq:psignal}
\end{equation}
Two properties shape what follows (Figure~\ref{fig:psignal}). Read against group size, coverage on a hard task is poor at the sizes prior work uses and good once the group is a few times larger: the same task flips from mostly silent to mostly informative with no change to the reward. Read against $p$, it collapses at both extremes---hard tasks early in training, mastered tasks late, when the signal runs out.

Worse, the groups that do carry signal on a hard task carry it badly: when a single rollout out of $n$ succeeds, standardization gives it an advantage of $\sqrt{n-1}$ while each failure receives only $-1/\sqrt{n-1}$ (appendix), so one lucky trajectory dominates the group's gradient. A signal that is silent most of the time and spiky when present is plausibly what published hard-data exclusions are observing~\citep{chen2025loop,gvpo2026}.

\paragraph{Fix: explore more.}
This motivates practices that restore coverage where it is scarcest. \emph{(1) Size the group from data, not a guess}: a pilot pass with the base policy estimates the hardest tier's success rate $\hat p_{\min}$, and for a target coverage $\tau$ the group size follows from Equation~\ref{eq:psignal},
\begin{equation}
n \;\gtrsim\; \frac{\ln(1-\tau)}{\ln(1-\hat p_{\min})}
\qquad (\psig \approx 1-(1-p)^n \text{ for small } p).
\label{eq:nsize}
\end{equation}
This is a first-order heuristic, not an exact prescription---it treats rollouts as independent and $\hat p_{\min}$ as a point estimate from a small pilot, so it should be read as a floor to size against the hardware budget rather than a value we claim optimal: if the affordable $n$ falls short, Equation~\ref{eq:nsize} at least says which tier stays starved. \emph{(2) Keep the hardest tasks}: we retain the full task distribution, hardest tier included---it is not intrinsically harmful, only starved, and becomes the last remaining source of gradient once easier tasks saturate (Section~\ref{sec:ablations}). \emph{(3) Uncapped per-turn generation}: environment returns are truncated to a fixed per-turn cap, but the policy's own generation is capped only by the total response budget, so the model learns its own allocation of thinking across turns---long, error-recovering episodes are exactly the ones a per-turn cap would truncate. None of these is a new algorithm; together they turn the hardest tier from poison into medicine, since the same data other work drops as harmful is, once explored enough, exactly where the sparse signal was missing.

\paragraph{Why not just densify the reward instead?}
Partial-credit rewards (Equation~\ref{eq:dense}) and step-level estimates~\citep{salt2026,agentevolver2025} manufacture within-group variance even in all-fail groups, which is why they look necessary when groups are small---but the substitute is imperfect, and may reward partial progress that still steers the trajectory wrong. Scaling exploration removes the reason to substitute (Section~\ref{sec:ablations}).

\subsection{Policy Drift, and Drift Less}
\label{sec:drift}

Explore-more manufactures the signal; it does not keep it trustworthy. We use \emph{policy drift} for the tendency of the sampling distribution---the one the policy actually rolls out from---to move away from the distribution the update implicitly assumes. This happens whenever an RL loop must extract many gradient steps from a small, repeatedly revisited task pool---the norm for interactive environments.

\paragraph{Diagnosis: four causes of drift.} We trace drift to four causes, numbered for reference through the rest of the section.
\emph{(1) An unanchored objective lets the sampling distribution narrow.} Repeatedly optimizing the same tasks reinforces high-reward token patterns and lets entropy fall; recent single-pass recipes profitably drop the KL penalty because they have plenty of fresh data to explore~\citep{yu2025dapo,liu2025drgrpo}. In our revisit-heavy regime that freedom is dangerous: dropping the anchor lets exploration collapse exactly when saturation is already making informative groups rare (Figure~\ref{fig:psignal}b), compounding Failure 1.
\emph{(2) Reusing rollouts across updates changes what the update sees.} Splitting a rollout batch into several mini-batch updates is the standard way to amortize rollout cost, and the importance ratio is designed to correct the resulting estimate. We simply avoid the question: the mini-batch is the whole batch, so no update consumes a sample it did not generate.
\emph{(3) Length-imbalanced loss averaging biases what little signal survives.} Averaging the loss per sequence before averaging across sequences divides each trajectory's contribution by its own length, down-weighting the long, error-recovering episodes a long-horizon agent most needs. This is drift, not merely lost signal: it steers the policy toward the short trajectories it already produces---the same narrowing as cause~(1), arriving through the loss denominator rather than the objective.
\emph{(4) Densified or substituted reward signals may carry their own error.} Partial-credit and step-level signals are imperfect proxies for task success; training on them pulls the policy toward the proxy rather than the goal, a distortion of the same family as (1)--(3) even though it originates in the reward rather than in sampling or loss.

\paragraph{Fix: drift less.}
Four choices answer the four causes. \method{} keeps the sampling and learning policies identical at every step---the gradient mini-batch is the whole rollout batch and a single pass is taken over it---removing cause (2). We use the sparse reward of Equation~\ref{eq:sparse} rather than the dense form, answering cause (4): a fully-correct-only signal has no proxy to drift toward. Causes (1) and (3) are answered inside the loss itself.

We keep the standard clipped form for notational continuity with GRPO and PPO, but the importance ratio and clip are inert here: one update per rollout batch with no stale reuse makes the ratio identically $1$ (verified in the logs; appendix). Let $\pi_{\theta_{\mathrm{old}}}$ denote the policy that generated the current batch, $M_{i,t}\in\{0,1\}$ mask environment tokens so gradient flows only through action tokens the policy controls, and $\mathcal F$ the set of fault-quarantined episodes (a serving-layer precondition detailed below). With the per-token ratio
\begin{equation}
\rho_{i,t}(\theta)=\frac{\pi_\theta(o_{i,t}\mid q,o_{i,<t})}{\pi_{\theta_{\mathrm{old}}}(o_{i,t}\mid q,o_{i,<t})},
\end{equation}
the clipped surrogate is
\begin{equation}
\mathcal{S}_{i,t} = \min\big(\rho_{i,t}\hat A_i,\ \mathrm{clip}(\rho_{i,t}, 1{-}\epsilon_{\mathrm{low}}, 1{+}\epsilon_{\mathrm{high}})\hat A_i\big),
\end{equation}
and the KL term $D_{\mathrm{KL}}(\pi_\theta,\pi_{\mathrm{ref}})_{i,t} \geq 0$ is estimated with the low-variance $k3$ estimator~\citep{schulman2020klapprox}, where $\pi_{\mathrm{ref}}$ is fixed to the base model throughout training. \method{} minimizes the token-mean loss obtained by \emph{adding} a KL penalty to the negative clipped surrogate,
\begin{equation}
\mathcal{L}(\theta) = \frac{1}{N} \sum_{i \notin \mathcal F} \sum_{t=1}^{|o_i|} M_{i,t} \Big[ -\mathcal{S}_{i,t} \;+\; \beta\, D_{\mathrm{KL}}(\pi_\theta,\pi_{\mathrm{ref}})_{i,t} \Big] ,
\label{eq:loss}
\end{equation}
with $N = \sum_{i\notin\mathcal F}\sum_t M_{i,t}$ the pooled action-token count and no entropy bonus. Two places differ from the original GRPO objective~\citep{shao2024deepseekmath}, both for drift. The single denominator $N$ pools every action token across the batch rather than normalizing per sequence first, answering cause (3): every action token counts equally regardless of trajectory length. The KL penalty answers cause (1), pulling the sampling distribution back toward the base model's breadth precisely when repeated optimization would narrow it. The mask $M_{i,t}$ confines both to tokens the policy emitted.

\paragraph{Environment reliability: fault quarantine.}
The policy executes arbitrary code, so an episode can end without a verdict for two reasons that must be told apart. \emph{Agent-induced} terminations---an infinite loop hitting the turn limit, the policy's own allocation exhausting memory---are genuine failures of the behavior under evaluation and are scored $0$ like any other. Only \emph{exogenous} faults the serving layer attributes to itself (a worker OOM-killed by a co-resident episode, a dead process) enter $\mathcal F$: under a binary reward they are indistinguishable from genuine failure and would inject a false negative into Equation~\ref{eq:adv}. Quarantine precedes scoring, so such an episode shrinks its group rather than contributing a zero. We also isolate concurrent episodes with bounded per-worker resources and recycle unhealthy workers; the appendix gives the full rule and its limits.

\subsection{Test-Time Budget Transfer}
\label{sec:transfer}

Training at a long interaction budget is costly and hard: every extra turn multiplies rollout time across the whole group, and the longer an episode runs the more ways it has to end in truncation or a fault rather than a verdict. We train at a moderate budget, sized to cover the successful hard-task trajectories of the pilot pass, and simply raise the turn count and context length at test time---no search, no multi-rollout selection. The payoff lands where the headroom is: the hardest tasks, and applications never seen in training (Section~\ref{sec:ablations}).

\section{Experiments}
\label{sec:experiments}

\subsection{Experimental Setup}
\label{sec:setup}

\paragraph{Benchmark and data.}
AppWorld~\citep{trivedi2024appworld} provides 9 applications, 457 APIs, $\sim$100 simulated users, and 735 tasks in four splits---Train 90 / Dev 60 / Test-Normal 168 / Test-Challenge 417---each judged by held-out, state-based unit tests; Test-Challenge also includes applications absent from training. TGC (task goal completion) is the fraction of tasks whose final state passes all tests; SGC the stricter fraction of scenarios whose three tasks all pass.

\paragraph{Implementation.}
We post-train Qwen3-14B~\citep{qwen3} with verl~\citep{sheng2024hybridflow}: asynchronous SGLang~\citep{zheng2023sglang} rollouts drive the multi-turn agent loop against our stabilized AppWorld server, and Megatron~\citep{shoeybi2019megatron} performs the updates. We set the rollout group size to $n{=}32$, within the range Equation~\ref{eq:nsize} suggests for the hardest retained tier at a moderate target coverage; Table~\ref{tab:config} gives the full configuration.
\begin{table}[!t]
\centering\footnotesize
\setlength{\tabcolsep}{0.8pt}
\begin{tabular}{lccccc}
\toprule
 & & \multicolumn{2}{c}{Test-Normal} & \multicolumn{2}{c}{Test-Challenge} \\
Method & Model & TGC & SGC & TGC & SGC \\
\midrule
\multicolumn{6}{l}{\emph{Trained-policy agents}}\\
\textbf{\method{} (ours)} & Qwen3-14B & \textbf{86.9} & \textbf{80.4} & \textbf{67.6} & \textbf{50.4} \\
ESAT$^{b}$ & Qwen3-14B & 75.2 & 63.6 & 58.5 & 47.5 \\
LOOP & Qwen2.5-32B & 72.6 & 53.6 & 47.2 & 28.8 \\
GVPO & Qwen2.5-32B & 72.6 & 55.4 & 49.4 & 28.8 \\
SAGE & Qwen2.5-32B & 72.0 & 60.7 & 50.1 & 32.4 \\
SALT$^{c}$ & Qwen2.5-32B & 66.2 & 47.9 & 36.8 & 20.9 \\
AgentEvolver$^{b}$ & Qwen2.5-14B & 48.7 & -- & -- & -- \\
ProST & Phi-4-14B & 46.4 & 28.6 & 17.8 & 8.6 \\
SGCD & Qwen3.5-4B & 45.6 & 17.9 & 27.0 & 8.5 \\
CuES & Qwen2.5-14B & 45.2 & -- & -- & -- \\
\midrule
\multicolumn{6}{l}{\emph{Training-free inference-time systems}}\\
HCL-GP$^\ddagger$ & Sonnet 4.6 & 98.2 & 98.2 & 98.3 & 97.8 \\
ASSAY & Sonnet 4.5 & 89.3 & 75.3 & -- & -- \\
ReAct$^{d}$ & Sonnet 4.5 & 83.9 & 70.3 & -- & -- \\
ACE & DeepSeek-V3.1 & 76.2 & 64.3 & 57.3 & 39.6 \\
CUGA & GPT-4.1 & 73.2 & 62.5 & 57.6 & 48.2 \\
SkillX$^{a}$ & GLM-4.6 & 64.9 & -- & -- & -- \\
METIS & GPT-4o & 60.1 & -- & -- & -- \\
ACON & GPT-4.1 & 56.5 & -- & -- & -- \\
ReMe$^{a}$ & Qwen3-14B & 42.0 & -- & -- & -- \\
\bottomrule
\end{tabular}
\caption{\textbf{AppWorld results} (official leaderboard~\citep{appworldleaderboard} or cited papers in \S\ref{sec:related}). Top: trained policies. Bottom: training-free systems around a fixed backbone, open or closed. All entries are mean@1 unless marked $^{a}$mean@4, $^{b}$mean@8, $^{c}$mean@3; $^{d}$reported by \citet{assay2026}; $^\ddagger$joint-scenario protocol, not comparable to per-task rows.}
\label{tab:main}
\end{table}

\begin{table}[!t]
\centering\footnotesize
\setlength{\tabcolsep}{2pt}
\begin{tabular}{@{}ll@{\;\,}ll@{}}
\toprule
\multicolumn{4}{l}{\emph{Training}} \\
base model & Qwen3-14B & learning rate & $3\times10^{-6}$ \\
tasks/steps/batch & 90/90/90 & KL $\beta$ / entropy & $10^{-4}$ / 0 \\
group size $n$ & 32 (2{,}880/step) & on-policy & 1 update/step \\
budget & 50 turns, 32k & temperature & 0.9 \\
prompt / obs.\ cap & 4k / 4k & checkpoint & step 90 (fixed) \\
per-turn gen.\ cap & none & hardest tier & kept \\
\midrule
\multicolumn{4}{l}{\emph{Inference (budget transfer)}} \\
budget & 100 turns, 61k & sampling & $T{=}0.6$, $p$ .95 \\
\bottomrule
\end{tabular}
\caption{Training and inference configuration.}
\label{tab:config}
\end{table}

\begin{table}[t]
\centering\footnotesize
\setlength{\tabcolsep}{3.2pt}
\begin{tabular}{@{}ll@{\;}cc@{\;\;}cc@{}}
\toprule
 & & \multicolumn{2}{c}{Test-Normal} & \multicolumn{2}{c}{Test-Challenge} \\
Policy & Budget & m@4 & b@4 & m@4 & b@4 \\
\midrule
Base & 100t/61k & 32.4 & 58.9 & 19.7 & 37.7 \\
\method{} & 50t/32k & 79.5 & 89.2 & 54.6 & 67.7 \\
\method{} & 100t/61k & 83.2 & 93.5 & 66.1 & 82.5 \\
\midrule
\multicolumn{2}{@{}l}{Leaderboard (m@1, 100t/61k)} & \multicolumn{2}{c}{\textbf{86.9}} & \multicolumn{2}{c}{\textbf{67.6}} \\
\bottomrule
\end{tabular}
\caption{\textbf{Metric map} (TGC, step-90 checkpoint vs.\ base; m@$k$ = mean@$k$, b@$k$ = best@$k$). Budget transfer, not a different model, closes the gap to the leaderboard entry.}
\label{tab:metrics}
\end{table}

\paragraph{Evaluation.}
We report mean@$k$ (average TGC over $k$ runs), best@$k$ (per-task union over $k$ runs), and the leaderboard submission (mean@1), under two inference budgets: training (50 turns / 32k) and enlarged (100 turns / 61k), mapped in Table~\ref{tab:metrics}. All evaluation uses the official AppWorld SDK and its unit tests, on the fixed step-90 checkpoint.

\subsection{Main Results}
\label{sec:main}

\begin{figure}[t]
\centering
\includegraphics[width=0.86\columnwidth]{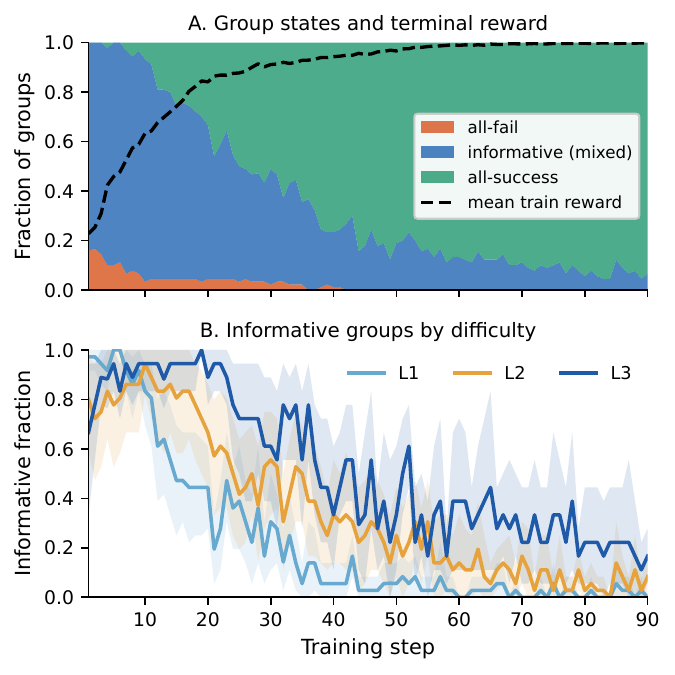}
\caption{\textbf{Training dynamics from rollout logs} ($n{=}32$, 90 steps; shaded: bootstrap CIs). \textbf{A}: group composition---convergence is the signal running out, the empirical face of Figure~\ref{fig:psignal}b. \textbf{B}: L3 stays informative far longer than L1/L2.}
\label{fig:dynamics}
\end{figure}

\begin{figure}[!t]
\centering
\includegraphics[width=0.98\columnwidth]{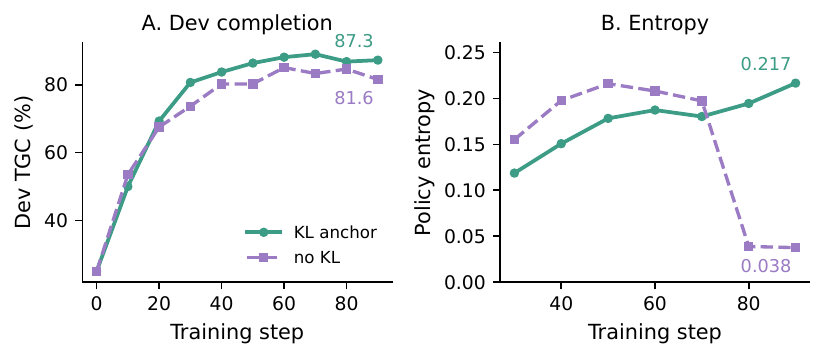}
\caption{\textbf{The KL anchor prevents late-phase collapse} (Dev TGC and policy entropy vs.\ training step). Past step $\sim$70 the unanchored run's entropy collapses (0.038) and Dev stalls at 81.6, while the anchored run stays healthy (0.217) and improves to 87.3.}
\label{fig:kl}
\end{figure}

\paragraph{A leaderboard-topping policy from simple but effective agentic RL.}
At submission (Feb.~2026) our single Qwen3-14B policy held the top of the AppWorld leaderboard---Test-Normal 86.9 TGC / 80.4 SGC and Test-Challenge 67.6 / 50.4---under the standard per-task protocol at that time (Table~\ref{tab:main}, Figure~\ref{fig:landscape}). It leads the next-best reported trained-policy result, ESAT, by nearly 12 TGC on Test-Normal and 9 on Test-Challenge, on one of the smallest backbones in that group, and it does so with the whole capability in the weights: at inference it is one checkpoint answering one prompt---no orchestration, skill library, retrieved memory, or test-time debugging. The gain is the training, not the backbone: on the same base at the same budget it adds more than 50 TGC points (Table~\ref{tab:metrics}).

Two systems post higher numbers, HCL-GP~\citep{hclgp2025} and ASSAY~\citep{assay2026}, and both buy them the same way: a frontier closed backbone many times our size, plus machinery around it---curated per-scenario skills, retrieval, and for HCL-GP a joint-scenario protocol with test-time debugging. That machinery has to be rebuilt for the next suite and leaves nothing behind in a set of weights; the same backbone as a plain ReAct agent lands in our policy's range (Table~\ref{tab:main}). \method{} needs none of it: the capability is in the model, and it is a model anyone can host.

\subsection{Analysis: Does the Diagnosis Hold Up?}
\label{sec:ablations}

\paragraph{The training dynamics match the coverage analysis.}
Figure~\ref{fig:dynamics} reads the mechanism off the rollout logs. All-fail groups disappear within $\sim$10 steps; as train reward saturates above 0.99, all-success groups take over and the informative fraction shrinks---the ``signal runs out'' regime Equation~\ref{eq:psignal} predicts at large $p$. Split by difficulty, the hardest tier stays informative long after easy tiers go silent: under adequate coverage L3 is the last gradient source, not noise.

\paragraph{The KL anchor keeps the distribution alive.}
Figure~\ref{fig:kl} isolates the primary drift cause. Anchored and unanchored runs learn near-identically for the first half; past step $\sim$70 the unanchored run's entropy collapses and its Dev score plateaus, while the anchored run's entropy holds and improvement continues to step 90 with no early stopping---drift made visible, exactly when saturation makes informative groups rarest.

\begin{figure}[t]
\centering
\includegraphics[width=0.88\columnwidth]{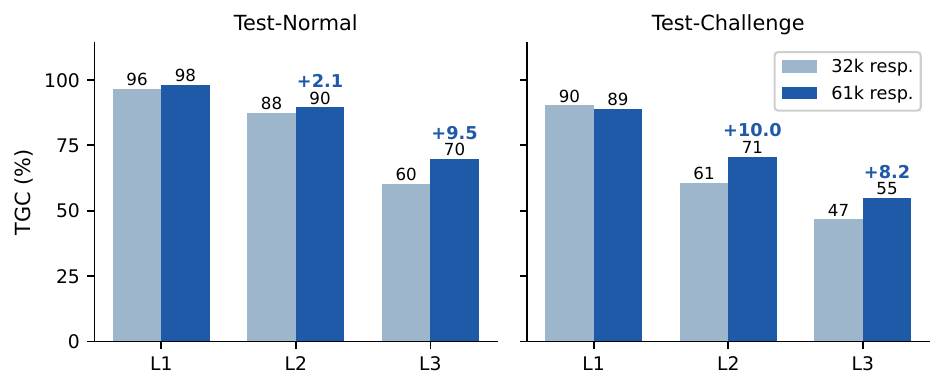}
\caption{\textbf{Budget transfer concentrates on the hardest tasks} (per-difficulty TGC, mean@1; 50t/32k to 100t/61k). Easy tiers are saturated; gains land on L3 and unseen applications.}
\label{fig:tts}
\end{figure}

\paragraph{Budget transfer works, and RL---not the budget---pays for it.}
Enlarging the budget lifts the trained policy from 79.5 to 83.2 mean@4 on Test-Normal (Table~\ref{tab:metrics}), the gain concentrated on the hardest tier and unseen applications (Figure~\ref{fig:tts}). It helps the base too, and by more (22.8 to 32.4), yet leaves it 47 points below what the trained policy reaches at the smaller budget: the budget is not what buys the capability. Against the finding that RL pass@1 $\le$ base pass@$k$~\citep{szot2026sge,yue2025limit}, our single run of 86.9 exceeds the base's best@4 of 58.9 by 28 points (67.6 vs.\ 37.7 on Test-Challenge)---at this scale RL adds capability resampling cannot reach.

\begin{figure}[t]
\centering
\includegraphics[width=0.88\columnwidth]{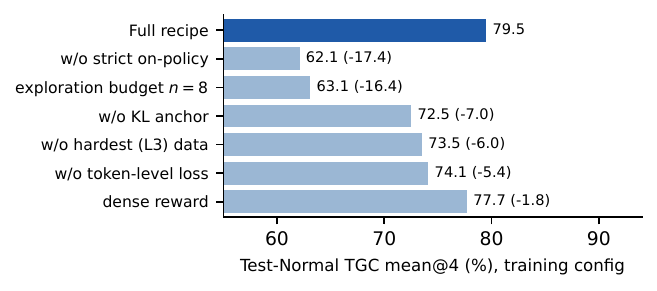}
\caption{\textbf{Component ablations} (Test-Normal TGC mean@4, training budget, step-90; one run per variant, retrained from scratch with one setting changed).}
\label{fig:ablation}
\end{figure}

\paragraph{Every component earns its place, and the ordering matches the theory.}
Each variant in Figure~\ref{fig:ablation} retrains 90 steps with one component changed from Table~\ref{tab:config}, and the ordering recovers the diagnosis. The two heaviest costs sit on opposite sides of it: on the coverage side, a group of $n{=}8$---the size prior work uses---rather than 32 costs $-16.4$, though at matched steps it gives up sampling along with coverage; on the drift side, setting the gradient mini-batch to half the rollout batch---each batch is then consumed in two successive updates, so the second update sees data drawn by the policy as it stood before the first---costs $-17.4$. The KL anchor is worth $-7.0$ and the token-level loss $-5.4$, both acting on the same narrowing through different routes; dropping the hardest tier costs $-6.0$, reversing the published finding at small $n$~\citep{chen2025loop,gvpo2026}. Densifying the reward moves the result least, $-1.8$: partial credit was a compensation for under-exploration rather than a necessity~\citep{chen2025loop,demystify2026}.

\subsection{Transfer to a Harder Domain: Real-Repository Software Repair}
\label{sec:swe}

\begin{table}[t]
\centering\footnotesize
\setlength{\tabcolsep}{6pt}
\begin{tabular}{lcc}
\toprule
Policy & mean@4 & best@4 \\
\midrule
Base (Qwen3.5-9B) & 31.3 & 43.8 \\
\method{} (training budget) & 47.9 & 58.0 \\
\method{} (enlarged budget) & 50.2 & 60.8 \\
\midrule
$\Delta$ (\method{} vs.\ base) & \textbf{+16.6} & \textbf{+14.2} \\
\bottomrule
\end{tabular}
\caption{\textbf{SWE-bench Verified} (resolve rate \%, Qwen3.5-9B, mini-swe-agent). Rows 1--2 use the 80-turn / 36k-token training budget; row 3 raises the turn and context budget at test time, as on AppWorld. $\Delta$ is row 2 vs.\ row 1, at matched budget.}
\label{tab:swe}
\end{table}

Software repair is the harder test of the same diagnosis: the agent works inside a real repository, drives a bash shell in a Docker container over many turns, and is judged only by whether its patch passes the project's own unit tests---a longer horizon, a larger action space, a task pool too big to memorize. We carry over the design principles, not the hyperparameter vector: sparse outcome reward, same-task groups sized for coverage, on-policy KL-anchored updates, token-level loss over action tokens. Three settings are re-tuned, named here so the transfer is not read as literal: $n{=}16$ rather than 32, since episodes cost far more; KL coefficient $10^{-2}$ rather than $10^{-4}$; and a constant $-0.2$ instead of $0$ for terminal states yielding no reviewable patch (crash, timeout, no patch, apply failure), separating ``produced nothing to test'' from ``produced a wrong patch''. The appendix lists every remaining difference. Training is on SWE-rebench~\citep{swerebench2025} with Qwen3.5-9B~\citep{qwen35} and mini-swe-agent~\citep{yang2024sweagent}, a purely bash-based harness with no repository-specific tooling; evaluation is on SWE-bench Verified~\citep{jimenez2024swebench,swebenchverified2024}. Against contamination we de-duplicate at the level of \emph{repositories} rather than instances, dropping every task from any repository appearing in Verified; the appendix gives the full configuration and filtering pipeline.

The principles transfer: \method{} lifts the resolve rate from 31.3 to 47.9 mean@4 and best@4 from 43.8 to 58.0, with budget transfer adding $+2.3$ (Table~\ref{tab:swe})---the same two failures, answered the same way, in another outcome-verified domain.

\section{Conclusion and Future Work}
\label{sec:limitations}
We examined whether the apparent limitations of outcome-only RL for long-horizon agents arise from sparse rewards alone or from how training is conducted. Our evidence points to two practical bottlenecks: obtaining informative outcome variation and limiting policy drift. \method{} addresses both with a simple recipe---explore more, drift less. At submission time, a single open 14B policy trained with this recipe reached the top of the AppWorld leaderboard, and the same principles improved software repair. For a well-defined domain, some of the capabilities that elaborate systems assemble at inference time can instead be internalized into a small open model's weights. Much recent progress has come from engineering around the model; our results suggest that the model itself remains a promising and comparatively underexplored direction.

Three directions follow. \emph{Environment scaling.} Our AppWorld training pool contains only 90 tasks in a standardized environment. Scaling to larger, more diverse, and more complex task distributions---including multilingual software engineering and harder benchmarks now appearing~\citep{swebenchpro2025,frontierswe2026,appworldul2026}---would test whether the gains persist and expose new limits. \emph{Better RL algorithms.} More sample-efficient methods could make that scale affordable while managing the exploration--exploitation trade-off more directly, for example by adapting group size or task sampling to the current success probability $p$. \emph{Domain mid-training.} Strengthening domain knowledge before RL could improve the base model's coverage and raise the attainable ceiling of outcome-based RL, complementing rather than replacing interaction-based post-training.

\clearpage
\bibliography{references}

\appendix
\section*{Technical Appendix}
\setcounter{section}{0}
\renewcommand{\thesection}{\Alph{section}}
\setcounter{table}{0}
\renewcommand{\thetable}{A\arabic{table}}
\setcounter{figure}{0}
\renewcommand{\thefigure}{A\arabic{figure}}

This appendix reports the configurations, log-level diagnostics, and cost
accounting behind the results in the main text. Every number below is read
off the training logs of the runs it describes; where a quantity was not
logged we say so rather than estimate it.

\section{Full AppWorld Training Configuration}
\label{app:hparams}

Table~\ref{tab:apphparams} gives the complete configuration of the main run.
Two entries deserve comment. First, \emph{train batch} and \emph{ppo
mini-batch} are both 90 \emph{tasks}: the rollout phase collects
$90\times32=2{,}880$ trajectories, and exactly one gradient step is taken on
all of them, so no rollout is ever reused (Appendix~\ref{app:onpolicy}).
Second, the micro-batch of 4 sequences per GPU is gradient accumulation only
--- it splits the backward pass, not the update --- and therefore introduces
no policy lag.

\begin{table*}[t]
\centering\footnotesize
\begin{tabular}{ll@{\qquad}ll}
\toprule
base model & Qwen3-14B & rollout group size $n$ & 32 \\
train tasks & 90 (hardest tier kept) & rollouts per step & 2{,}880 \\
train batch / mini-batch & 90 / 90 tasks, 1 epoch & validation group size & 4 \\
optimizer / LR & AdamW / $3\times10^{-6}$, constant & turns / response tokens & 50 / 32{,}768 \\
clip $\epsilon$ (low/high) & 0.2 / 0.2 & prompt / observation cap & 4{,}096 / 4{,}096 \\
KL coef $\beta$ / reference & $10^{-4}$ ($k3$) / fixed base & per-turn generation cap & none \\
entropy bonus & 0 & advantage estimator & GRPO, std-normalized \\
micro-batch / GPU & 4 (accumulation only) & reward & sparse $\{0,1\}$, all unit tests \\
hardware & 8 nodes $\times$ 8 GPUs & loss aggregation & token-mean, pooled \\
parallelism & Megatron TP8, PP1, CP2 & sampling temperature (train) & 0.9 \\
rollout engine & SGLang, async, TP8 & inference budget & 100 turns / 61k (YaRN) \\
checkpoint & step 90 (fixed) & inference sampling & $T{=}0.6$, top-$k$ 20, top-$p$ 0.95 \\
\bottomrule
\end{tabular}
\caption{Full AppWorld training configuration (main run, 90 steps).}
\label{tab:apphparams}
\end{table*}

\section{Advantage Magnitudes in a Sparse, Nearly-Degenerate Group}
\label{app:advmag}

Section~\ref{sec:starvation} states that a lone success in a group of $n$
receives a standardized advantage of $\sqrt{n-1}$. The derivation is one line.
Under the sparse reward, a group with $k$ successes out of $n$ has empirical
mean $\bar r = k/n$ and population standard deviation
$\sigma=\sqrt{\bar r(1-\bar r)}=\sqrt{k(n-k)}\,/\,n$. Substituting into the
standardized advantage of Equation~\ref{eq:adv}, a successful rollout
receives
\[
\hat A_+ = \frac{1-k/n}{\sigma} = \sqrt{\frac{n-k}{k}},
\qquad
\hat A_- = -\sqrt{\frac{k}{n-k}} ,
\]
and at $k=1$ this is $\hat A_+=\sqrt{n-1}$ against $\hat A_-=-1/\sqrt{n-1}$: the
single lucky trajectory is weighted $n-1$ times more heavily than each of the
$n-1$ failures, and carries essentially the whole group's gradient. The
estimator is not biased by this --- the advantages still sum to zero --- but the
per-step gradient from a hard task becomes a coin flip on whether that one
rollout appeared, which is the spiky signal small groups deliver at low $p$.
Enlarging $n$ does not remove the $k=1$ case; it makes it a small fraction of
the informative groups instead of nearly all of them.

\section{Verifying That Updates Are Strictly On-Policy}
\label{app:onpolicy}

``Strictly on-policy'' is a property we can check in the logs rather than merely
assert. When the gradient mini-batch equals the rollout batch and a single pass
is taken over it, the sampling and learning policies are the same network, so
the per-token importance ratio $\rho_{i,t}$ is exactly $1$, the PPO clip is
never active, and the measured policy KL between sampling and learning
distributions is exactly $0$.

Both quantities are logged every step, and over all 90 steps of the main run
\texttt{actor/pg\_clipfrac} and \texttt{actor/ppo\_kl} are both $0$ --- not
small, but identically zero at every step, in both clip directions. The clipped
surrogate therefore reduces to the plain policy gradient throughout training,
and the objective in the main text is the objective that actually ran. For
contrast, the variant of Appendix~\ref{app:offpolicy} logs a nonzero clip
fraction from its very first step.

\section{The ``w/o Strict On-Policy'' Variant}
\label{app:offpolicy}

This variant is the mildest departure from strict on-policy we could construct.
It runs the same training script with one change: the gradient mini-batch is
halved, so each rollout batch is consumed in \emph{two} sequential updates
instead of one. There is no replay buffer, no reuse of rollouts across
iterations, and no asynchronous generation. Concretely, the run used a train
batch of 88 tasks with a mini-batch of 44 (2 updates per rollout phase) against
the main run's 90/90; the 88-task batch is an artifact of that run's
configuration, and the two-task difference in pool coverage is far too small to
account for the gap below. The variant also sets \texttt{loss\_agg\_mode} to
\texttt{token-mean}, matching the main recipe. Step count, learning rate, group
size, KL coefficient, horizon, and reward are unchanged; rollouts per step
follow the batch ($88\times32=2{,}816$ against $2{,}880$).

\paragraph{The variant is not a single-variable contrast.}
Halving the mini-batch changes more than the policy lag, and the $-17.4$ point
gap cannot be attributed to lag alone. Within a fixed rollout batch it also
doubles the number of optimizer updates (180 against 90 over training), halves
the number of trajectories each gradient estimate averages over and so raises
its variance, and changes the trajectory of the Adam moment estimates and of
the effective step size. Any of these can move the endpoint on its own. The
honest reading is that this variant prices \emph{the whole package} of
consuming a rollout batch in two half-batch updates instead of one full-batch
update --- which is what a practitioner actually chooses between --- and that
it does not isolate policy lag as the mechanism. The main text accordingly
reports what the variant does and what it costs, without attributing the cost
to lag. Separating the factors would need at least a same-mini-batch,
double-update control and a matched-update-count control, neither of which we
ran.

What we can say about the lag itself is that the importance ratio is never the
binding constraint: over 90 steps the clip fraction averages
$1.5\times10^{-3}$ and peaks at $2.7\times10^{-3}$, the measured policy KL
averages $1.6\times10^{-4}$, and under 0.3\% of tokens are ever clipped. So
whatever produced the gap, it is not the surrogate being throttled by stale
ratios.

What does change is the sampling distribution. At step 90 the variant's policy
entropy has fallen to 0.036, against 0.217 for the main run --- a collapse
quantitatively indistinguishable from what removing the KL anchor produces
(0.038, Figure~\ref{fig:kl}), and reached by a different route. The
resulting Test-Normal score is 62.1 mean@4 against 79.5.
Table~\ref{tab:variants} places this alongside the other variants.

\section{Per-Variant Step-90 Diagnostics}
\label{app:variants}

Table~\ref{tab:variants} reports, for every ablation, the internal training
diagnostics at the step-90 checkpoint next to the scores. Three patterns are
worth noting. (i) Final entropy orders the variants almost exactly as final
score does, which is the drift account in one column. (ii) The no-KL run's
logged gradient norm reaches $4.7\times10^{4}$ at step 90, four orders of
magnitude above every anchored run --- the entropy collapse is accompanied by
outright optimization instability, not a quiet plateau. (iii) Train reward is
saturated above 0.93 for all variants, so the score differences are not
explained by any of them failing to fit the training pool; they differ in what
distribution they arrive at.

\begin{table*}[t]
\centering\footnotesize
\begin{tabular}{lccccccc}
\toprule
Variant & entropy & clip frac & grad norm & train reward & Dev & Test-Normal & TN $\sigma$ \\
\midrule
Full recipe & 0.217 & $0$ & 0.014 & 0.997 & 87.3 & \textbf{79.5} & 0.132 \\
dense reward & 0.194 & $0$ & 0.018 & 0.999 & 86.0 & 77.7 & 0.112 \\
w/o token-level loss & 0.140 & $0$ & 0.019 & 0.994 & 85.1 & 74.1 & 0.163 \\
w/o hardest (L3) data & 0.093 & $0$ & 0.037 & 0.993 & 83.8 & 73.5 & 0.125 \\
w/o KL anchor & 0.038 & $0$ & $4.7\times10^{4}$ & 0.984 & 81.6 & 72.5 & 0.153 \\
exploration budget $n{=}8$ & 0.156 & $0$ & 0.045 & 0.931 & 74.1 & 63.1 & 0.160 \\
w/o strict on-policy & 0.036 & $9\times10^{-4}$ & 0.064 & 0.946 & 71.1 & 62.1 & 0.187 \\
\bottomrule
\end{tabular}
\caption{Per-variant diagnostics at the step-90 checkpoint. Entropy, clip frac,
grad norm, and train reward are training-log quantities at that step, so the
off-policy clip fraction is its step-90 value, not the $1.5\times10^{-3}$
training average of Appendix~\ref{app:offpolicy}; Dev and Test-Normal
are TGC mean@4 at the training budget (50 turns / 32k). ``$\sigma$'' is the
across-run standard deviation of per-task reward over the 4 inference runs,
averaged over tasks. One training run per variant.}
\label{tab:variants}
\end{table*}

\paragraph{On the missing variance estimate.}
We report one training run per variant; independent training seeds for seven
90-step runs were beyond our compute budget, and we do not claim the ablation
deltas are separated at any particular confidence level. What we can bound is
the \emph{evaluation} contribution to the noise. The $\sigma$ column of
Table~\ref{tab:variants} is a per-task standard deviation on the $[0,1]$ reward
scale, so the standard error of a 4-run mean over the 168 Test-Normal tasks is
about $100\cdot0.13/\sqrt{4\cdot168}\approx0.5$ TGC points: differences of a
point or two are near the edge of inference noise alone, while the $-16$ and
$-17$ entries are far outside it. The dense-reward gap of
$-1.8$ is the one we would treat as suggestive rather than established; its
interpretation --- that partial credit is
not \emph{necessary} once exploration is adequate --- rests on the gap being
small, which is the robust direction to be wrong in.

\section{Compute and Sampling Cost}
\label{app:cost}

``Minimalist'' in this paper refers to the number of moving parts, not to
compute frugality, and the accounting below is deliberately explicit about
that. The main AppWorld run consumed:

\begin{itemize}\itemsep1pt
\item \textbf{259{,}200 trajectories} ($90$ steps $\times\,90$ tasks
$\times\,n{=}32$).
\item \textbf{$3.47\times10^{9}$ tokens} processed across generation and
training, summed over steps from the logged per-step token counts.
\item \textbf{42.0 hours wall-clock} on 8 nodes $\times$ 8 GPUs,
i.e.\ $\approx$2{,}690 GPU-hours; mean 1{,}680\,s per step, of which
generation is the dominant term early in training and shrinks as trajectories
shorten.
\item Mean trajectory length 10{,}979 response tokens over 42.5 assistant
turns; mean prompt 2{,}417 tokens. The 32k response cap binds on 0.18\% of
trajectories on average (max 2.5\% in any step), so the budget is not
silently truncating the distribution we train on.
\item Actor MFU averaged 0.80.
\end{itemize}

The $n{=}8$ ablation therefore trains on one quarter of the trajectories at
matched steps, so its $-16.4$ bounds the coverage effect from above rather than
isolating it, as Section~\ref{sec:ablations} notes: a compute-matched
comparison (e.g.\ $n{=}8$ with a $4\times$ larger task batch, or $n{=}8$ for
$4\times$ the steps) is the experiment that would separate coverage from total
sampling, and we did not run it. We flag this as the most important open check
on our central claim.

\section{Environment Reliability and Fault Quarantine}
\label{app:env}

The training-side environment is a multi-node, multi-process server--client
architecture over the official AppWorld engine: per-worker memory caps with
automatic recycling, request timeouts with worker repair, and episode-level
state isolation.

\paragraph{What is and is not quarantined.}
The distinction is between \emph{agent-induced} and \emph{exogenous}
terminations, because the policy executes arbitrary code and can itself cause
every failure mode the infrastructure can. Agent-induced cases are scored as
failures ($r=0$) exactly like a wrong answer: a rollout that times out because
the policy wrote an infinite loop or a pathological API-call pattern; an
episode whose own allocation exhausts its memory budget; a sequence of requests
that corrupts only that episode's state. A quarantine flag is raised only when
the serving layer attributes the fault to itself and not to the trajectory ---
a worker OOM-killed while a co-resident episode was the allocator, a
non-responsive or crashed worker process, state corruption that crosses
episode boundaries, an infrastructure-level request failure. Quarantined
episodes are dropped from the loss via $\mathcal F$ rather than scored as
failures, because under a binary reward a server-caused crash and a genuine
failure are otherwise indistinguishable to the estimator.

\paragraph{Ordering relative to the advantage.}
Quarantine is applied before scoring, not after: the surviving rollouts of a
group are what Equation~\ref{eq:adv} standardizes over, so
$\mathrm{mean}$ and $\mathrm{std}$ are computed on the reduced group and a
quarantined episode contributes neither a reward nor a token to the loss
denominator $N$. A group reduced to a single outcome class produces zero
advantage and no gradient, as any degenerate group does. The alternative
ordering --- score first, then mask the loss --- would leave the false negative
in the group statistics, which is what we want to avoid.

\paragraph{Three caveats for reproduction.}
First, the attribution is a serving-layer heuristic, not a proof. A fault the
policy caused but the server attributes to itself would be removed from the
group and bias it optimistically; a fault the server caused but attributes to
the trajectory is scored as a failure, in the conservative direction. We know of
no way to make this exact for an agent running arbitrary code inside the
environment's own process tree.

Second, we did not log a per-step quarantine count, so we cannot report an
exclusion rate or the sensitivity experiment a reader should reasonably want:
retraining with every quarantined episode scored as a failure, which would
upper-bound the selection effect. This is a gap in our instrumentation, and it
is the second experiment we would add after the compute-matched group-size run
of Appendix~\ref{app:cost}. The one diagnostic we do have is that the logged
aborted-response ratio was 0 for all 90 steps of the main run, i.e.\ no
trajectory was lost to generation-side aborts.

Third, the effect of this layer is large and easy to underestimate: before it
existed, otherwise-similar runs plateaued near 47 TGC through silent
environment corruption, and comparable recipes reached 73--74 after
stabilization under the same evaluation. Since we cannot decompose that shift
into ``fewer false negatives'' versus ``fewer lost episodes'', it should be read
as evidence that the serving layer matters at a scale comparable to the
algorithmic choices, and as a reason to treat the quarantine rule as part of the
protocol to be reproduced rather than an implementation detail. Reproducing
these numbers on an unstabilized server should be expected to yield the former,
not the latter.

\begin{table*}[t]
\centering\footnotesize
\begin{tabular}{ll@{\qquad}ll}
\toprule
base model & Qwen3.5-9B & rollout group size $n$ & 16 \\
train tasks & 5{,}639 (1{,}667 repos) & validation group size & 4 \\
train batch / ppo mini-batch & 64 / 64 tasks & max assistant turns & 80 \\
ppo epochs & 1 (strict on-policy) & max response tokens & 36{,}864 \\
steps & 260 ($\approx$3 epochs) & max prompt tokens & 4{,}096 \\
optimizer LR & $4\times10^{-6}$ & tool response cap & 4{,}096 \\
clip $\epsilon$ (low/high) & 0.2 / 0.2 & harness & mini-swe-agent (bash only) \\
KL coef $\beta$ & $10^{-2}$ ($k3$) & thinking mode & disabled \\
entropy bonus & 0 & reward & sparse $\{0,1\}$ on test outcome \\
micro-batch / GPU & 1 & no-patch / crash / timeout / apply-fail & $-0.2$ each \\
sampling temperature & 1.0 & rollout timeout & 5{,}400\,s \\
hardware & 8 nodes $\times$ 8 GPUs & env budget / episode & 2 CPU, 6\,GB \\
parallelism & Megatron TP4, PP2, CP1 & rollout engine & SGLang, TP2 \\
\bottomrule
\end{tabular}
\caption{Full SWE-bench training configuration.}
\label{tab:swehparams}
\end{table*}

\section{SWE-bench Configuration and Data Construction}
\label{app:swe}

\paragraph{Training data.}
We start from SWE-rebench~\citep{swerebench2025} and apply three filters, in
order: (i) the task's Docker image must be locally materializable, (ii)
\emph{repository-level} de-duplication against SWE-bench Verified --- we
collect the set of repositories appearing anywhere in Verified and drop every
SWE-rebench instance from any of them, rather than merely dropping matching
instance IDs, and (iii) prompt length $\le$ 4{,}096 tokens. The result is
5{,}639 tasks spanning 1{,}667 distinct repositories, mean prompt length
1{,}821 tokens. Filter (ii) is the strict choice: it removes not only the
evaluation instances but also every other issue from the same codebases, so no
training task shares a repository with any evaluation task. This costs training
data on exactly the popular repositories that would help most, which is the
trade we want for a contamination claim.

\paragraph{Evaluation.}
SWE-bench Verified~\citep{jimenez2024swebench,swebenchverified2024}, all 500
instances, mini-swe-agent~\citep{yang2024sweagent} as the harness (a purely
bash-based scaffold with no repository-specific tooling). Five instances whose
Docker images we could not obtain are scored 0 rather than excluded, so the
reported resolve rates are computed over the full 500 and are, by that amount,
conservative.

\paragraph{Configuration.}
Table~\ref{tab:swehparams} gives the full setup. The design principles are the
same as on AppWorld --- strictly on-policy (batch $=$ mini-batch $=$ 64 tasks,
one update per rollout phase), outcome-only sparse reward, KL-anchored to the
base model, token-level pooled loss --- but the hyperparameters are not
identical, and we list every difference rather than describe the transfer as
literal: a smaller rollout group ($n{=}16$ against 32, since each episode is
far more expensive here, so coverage is bought at a lower target), a larger KL
coefficient ($10^{-2}$ against $10^{-4}$), a longer horizon (80 turns / 36k
response tokens against 50 / 32k), a different base model and optimizer LR
($4\times10^{-6}$ against $3\times10^{-6}$), thinking mode disabled, sampling
temperature 1.0 against 0.9, and a small set of negative constants for terminal
states that produce no reviewable patch (crash, timeout, no patch emitted,
patch fails to apply: $-0.2$ each) where AppWorld used $\{0,1\}$ throughout.
The claim the SWE-bench result supports is therefore that the two diagnoses and
the four practices answering them carry to another outcome-verified domain, not
that a fixed hyperparameter vector does. Instances that reach evaluation but fail their tests
score 0, so the negative constants separate ``produced nothing to test'' from
``produced a wrong patch'' without introducing partial credit for the patch
content itself.

\section{Evaluation Protocol}
\label{app:protocols}

All AppWorld evaluations use the fixed step-90 checkpoint. We keep three
configurations separate within every table and figure:
(i) \emph{training config} --- 50 turns / 32k response, training sampling
($T{=}0.9$), mean@4; used for all ablations.
(ii) \emph{scaled config} --- 100 turns / 61k via YaRN~\citep{peng2023yarn},
base-recommended sampling ($T{=}0.6$, top-$k$ 20, top-$p$ 0.95), mean@4.
(iii) \emph{leaderboard} --- scaled config, mean@1, official AppWorld harness
and unit tests.

\section{Scope of the Claims}
\label{app:scope}

Both of our domains are code-executing environments with automatic, state- or
test-based verifiers, and both reward functions are computed by running code.
The evidence therefore supports the claim that outcome-only RL suffices for
long-horizon interactive agents \emph{of this kind} --- an interpreter or a
shell, a programmatic verifier, tens of turns. It does not establish the same
for browser and GUI control, open-ended research, or any task whose success
cannot be checked automatically; there the signal-starvation analysis still
applies formally, but the ``manufacture the signal'' remedy depends on a
verifier we would not have. Extending the protocol to partially verifiable
domains is open.

\section{Reproducibility}
\label{app:repro}

All evaluation uses the official AppWorld and SWE-bench harnesses and their held-out
unit tests.

\end{document}